%% file: PaperForReview.tex
\documentclass[10pt,twocolumn,letterpaper]{article}

\usepackage[pagenumbers]{wacv} 

\usepackage{graphicx}
\usepackage{amsmath}
\usepackage{amssymb}
\usepackage{booktabs}

\usepackage{algorithm}
\usepackage[noend]{algpseudocode}
\usepackage{array}
\usepackage{tabulary}
\usepackage{graphicx}
\usepackage{xcolor}
\usepackage{bbm}
\usepackage{bm}
\usepackage[accsupp]{axessibility}  
\usepackage{float}
\usepackage{stfloats}
\usepackage{subfiles}

\usepackage[pagebackref,breaklinks,colorlinks]{hyperref}

\usepackage[capitalize]{cleveref}
\crefname{section}{Sec.}{Secs.}
\Crefname{section}{Section}{Sections}
\Crefname{table}{Table}{Tables}
\crefname{table}{Tab.}{Tabs.}

\def\wacvPaperID{76} 
\def\confName{WACV}
\def\confYear{2024}

\begin{document}

\title{Continual Test-time Domain Adaptation via Dynamic Sample Selection}

\author{ Yanshuo Wang$^{1,2}$, Jie Hong$^{1,2}$, Ali Cheraghian$^{1,2}$, Shafin Rahman$^{3}$\\ David Ahmedt-Aristizabal$^{2}$,
Lars Petersson$^{2}$, Mehrtash Harandi$^{2,4}$ \\
$^{1}$Australian National University, $^{2}$Data61-CSIRO, Australia\\ $^{3}$ North South University, Bangladesh, $^{4}$Monash University, Australia\\
{\tt\small \{Yanshuo.Wang,~ Ali.Cheraghian,~david.ahmedtaristizabal,~Lars.Petersson\}@data61.csiro.au,}\\
{\tt\small shafin.rahman@northsouth.edu,jie.hong@anu.edu.au, mehrtash.harandi@monash.edu}
}


\maketitle

\begin{abstract}
The objective of Continual Test-time Domain Adaptation (CTDA) is to gradually adapt a pre-trained model to a sequence of target domains without accessing the source data. This paper proposes a Dynamic Sample Selection (DSS) method for CTDA. DSS consists of dynamic thresholding, positive learning, and negative learning processes. Traditionally, models learn from unlabeled unknown environment data and equally rely on all samples' pseudo-labels to update their parameters through self-training. However, noisy predictions exist in these pseudo-labels, so all samples are not equally trustworthy. Therefore, in our method, a dynamic thresholding module is first designed to select suspected low-quality from high-quality samples. The selected low-quality samples are more likely to be wrongly predicted. 
Therefore, we apply joint positive and negative learning on both high- and low-quality samples to reduce the risk of using wrong information.
We conduct extensive experiments that demonstrate the effectiveness of our proposed method for CTDA in the image domain, outperforming the state-of-the-art results. Furthermore, our approach is also evaluated in the 3D point cloud domain, showcasing its versatility and potential for broader applicability.
\end{abstract}

\section{Introduction}
\label{sec:intro}
Consider an intelligent agent moving around non-stationary environments 
where the input domain gradually changes over time. As an illustration, a self-driving car could transition from daylight to darkness and then enter a snowy scenario.
Another example is an automated robot processing images captured from multiple continual domains like blurry and bright domains.
Therefore, the intelligent agent requires \textit{Continual Test-Time domain Adaptation} (CTDA) to adapt to gradually changing environments. 
Because of its real-life applicability, researchers have recently started exploring this new area~\cite{wang2022continual,dobler2022robust}. In contrast to traditional domain adaptation tasks, CTDA imposes some constraints: 
1) There should be multiple continual test domains instead of one, and the model should gradually adapt to continual domains using unlabeled test data.
2) The source data with which the model is trained cannot be used during the adaptation process. Instead, it is only allowed to leverage the pre-trained model developed from the source data.

\begin{figure}[t]\centering
\includegraphics[width=1.0\linewidth]{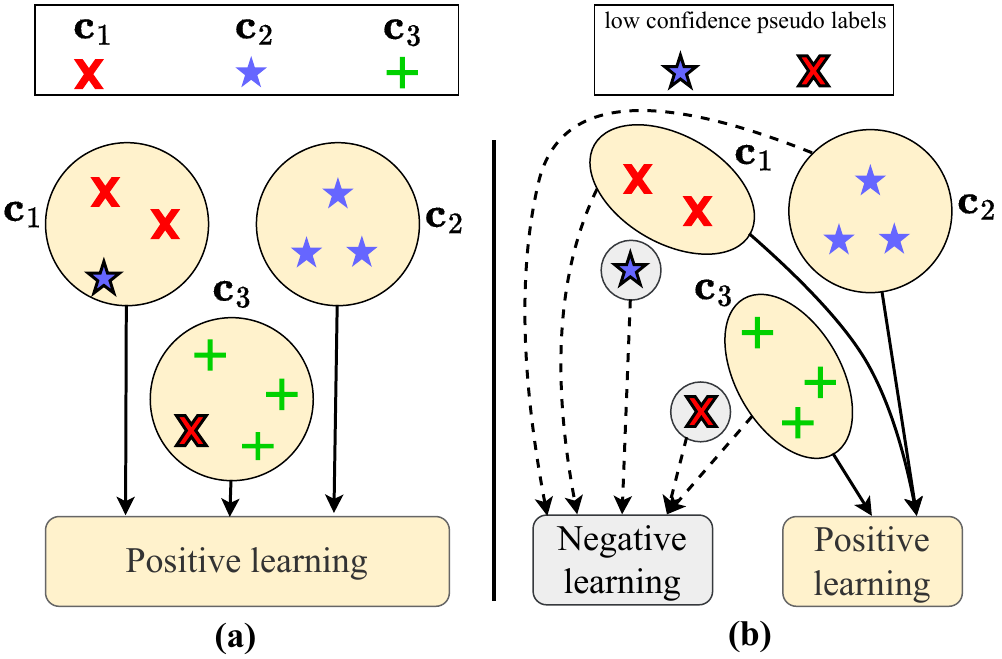}
\caption{Illustration of our learning strategy: 
\textbf{(a)} Pseudo labeling described is accomplished by the mean-teacher model, where a teacher model yields a pseudo label for a given sample. Some of the unlabeled samples probably assign incorrect pseudo-labels. Hence, treating all samples equally leads to error accumulation. \textbf{(b)} Our proposed method introduces Dynamic Sample Selection (DSS), to monitor the pseudo label generated by the mean teacher model and it selects low-quality pseudo labels via our proposed dynamic thresholding process. Ultimately, we use positive learning for samples with high-quality pseudo-label and negative learning for all samples.}
\label{fig:architecture_intro}
\end{figure}

Error accumulation is a well-known challenge in CTDA. In this task, the model adapts continual domains during test time on unlabeled test samples. The model normally relies on samples' pseudo-labels for the adaptation. However, not all pseudo-labels are reliable. As the erroneous pseudo-labels arise from the continual domains, the network tends to absorb inaccurate feedback, leading to an incorrect adjustment of its parameters over time. This accumulation of errors from the previous domains would have an adverse impact on the model's performance in the subsequent domains. To address this, the existing SOTA method~\cite{wang2022continual} generates augmented samples of a given input sample and calculates an augmentation-averaged prediction score from a teacher model. The assumption is that pseudo-labels calculated from the augmentation-averaged prediction score become more accurate over time. 

To address the aforementioned problem, we introduce a Dynamic Sample Selection (DSS) method, which incorporates jointly dynamic thresholding, positive, and negative learning. More specifically, to mitigate the issue of error accumulation, using dynamic thresholding, we design a class-wise confidence metric to evaluate the prediction quality of the pseudo-label for each sample in a batch. Dynamic thresholding monitors the confidence and identifies high- and low-quality samples. As shown in Figure \ref{fig:architecture_intro}, our proposed method treats the unlabeled samples differently based on confidence. We employ only the samples with higher confidence for positive learning, thus reducing the amount of noise introduced. At the same time, we utilize all examples for negative learning. The negative learning aims to eliminate the impact of low-confidence class prediction. A prediction that is potentially considered wrong is replaced by a negative complementary label. Traditionally, CTDA methods are tested on only 2D image datasets. However, in addition to challenging 2D image datasets (CIFAR-10-C~\cite{hendrycks2018benchmarking}, CIFAR-100-C~\cite{hendrycks2018benchmarking}, and ImageNet-C~\cite{hendrycks2018benchmarking}), we  experimented with 3D point cloud data ScanObjectNN-C~\cite{sun2022benchmarking}), demonstrating the robustness of our proposed method. Our method consistently outperforms existing methods by a decent margin.

In summary, contributions of this work are three-fold:
\begin{itemize}
\item We develop a novel method, Dynamic Sample Selection (DSS) for the CTDA task. DSS selects high- and low-quality samples for training, and it effectively reduces the negative impact of error accumulation of CTDA.
\item We apply joint positive and negative learning on noisy pseudo-labels to reduce the risk of propagating misleading information during the CTDA task.
\item Our study is pioneering in investigating the CTDA problem in the domain of 3D point cloud objects. We have benchmarked the performance of existing 2D approaches on 3D data and identified new challenges associated with 3D point cloud objects.
\end{itemize}

\section{Related work}
\noindent\textbf{Test-time Domain Adaptation.}
Compared with traditional unsupervised domain adaptation, Test-Time domain Adaptation (TTA) adapts the model trained from the source domain to the novel target domain without access to the original source data during inference time. One popular approach to reducing the domain gap in the absence of source data is to fine-tune the source model by adopting an unsupervised loss function based on the target distribution. 
Test entropy minimization (TENT)~\cite{wang2021tent} updates the trainable batch normalization parameters from a pre-trained model at test time by minimizing the entropy of the model prediction. Source hypothesis transfer (SHOT)~\cite{liang2020we} employs an entropy minimization and diversity regularizer with label smoothing techniques to train a general feature extractor from a pre-trained source model. Alternatively, Test-Time Training (TTT)~\cite{sun2020test} introduced a self-supervised rotation prediction auxiliary task to update model parameters for novel target samples. Test-Time Training with Masked Autoencoders~\cite{gandelsman2022test} similarly adopts the idea of masked autoencoders as an auxiliary task to train the model for each test sample. Recently AdaContrast\cite{chen2022contrastive} first leveraged contrastive learning with online pseudo refinement to learn better feature representations with less noisy pseudo labels. Conjugate PL~\cite{goyal2022test} propose a general way of obtaining test-time adaptation loss which is used for more robust predictions under distribution shifts.

\noindent\textbf{Continual Test-time Domain Adaptation.}
In addressing the challenge of Continual Test-time Domain Adaptation (CTDA), various solutions have been proposed. Notably, the online version of Tent, introduced by Wang~\textit{et al.}~\cite{wang2021tent}, presents an applicable approach. However, a drawback of earlier methods lies in their reliance on source data during inference. A significant advancement in this field is the Continual Test-time Adaptation Approach (CoTTA) developed by Wang~\textit{et al.}~\cite{wang2022continual}. CoTTA is notably the first method explicitly tailored to the demands of online continual test-time adaptation. This method employs a weighted augmentation-averaged mean teacher framework, building upon the insights from prior work such as the mean teacher predictions introduced by Tarvainen and Valpola~\cite{NIPS2017_68053af2}. Remarkably, the student-teacher framework proposed by CoTTA~\cite{wang2022continual} serves as a foundational architecture for numerous subsequent studies. In particular, Niu ~\textit{et al.}~\cite{niu2022efficient}, adopt a similar student-teacher framework. They incorporate continuous batch normalization statistics updates to reduce computational costs, thereby refining the efficiency of the adaptation process. Another notable avenue explored by researchers is the utilization of a mean teacher setup with symmetric cross-entropy and contrastive learning, as demonstrated in the work by Dobler ~\textit{et al.}~\cite{dobler2022robust} under the name RMT. While this approach introduces valuable insights, it remains dependent on source data to establish the source prototypes during a warm-up stage. Consequently, RMT cannot be categorized as a truly source-free approach within the realm of continual test-time domain adaptation. Alternatively, Gan ~\textit{et al.}~\cite{gan2022decorate} present an innovative strategy that leverages visual prompt learning in conjunction with a homeostasis-based adapting strategy.


\noindent\textbf{Pseudo Labelling.}
Pseudo-labelling is a widely used technique in semi-supervised learning or self-learning. It uses the model’s output class probability as a label for training. FixMatch\cite{sohn2020fixmatch} is a semi-supervised method that generates pseudo labels using the model’s predictions on weakly augmented unlabeled images. Then the model is trained to match the pseudo-label with the prediction on strongly-augmented images.
In contrast, UPS~\cite{rizve2021defense} not only produces 
reliable pseudo-labels with high confidence and low uncertainty, but it also incorporates negative learning to further reduce the model calibration error.
In the context of unsupervised domain adaptation, the process of pseudo-labeling frequently involves generating labels 
for unlabeled target samples by utilizing the predicted class probability of the source model. Existing methods~\cite{wang2022continual,dobler2022robust} mainly rely on pseudo labels as a form of ``supervision'' in order to compensate for the absence of ground truth labels in the target domain. 
However, they did not closely investigate the quality of pseudo labels, as mislabeled samples used in self-learning accelerate error accumulation ultimately. In contrast, in this paper, we propose joint positive and negative learning with dynamic threshold modules to minimize the effect of error accumulation from mislabeled pseudo labels.

\section{Dynamic Sample Selection}
\subsection{Problem Formulation}
Given a sequence of domains $\mathcal{D} = \{ \mathcal{D}_{1}$, \ldots, $\mathcal{D}_{t}$, \ldots, $\mathcal{D}_{T}$\}, 
we define the domain at time step $t$ as $\mathcal{D}_{t} = \{\textbf{x}_{t,i}\}_{i=1}^{n_{t}}$, where $\textbf{x}_{t,i}$ is the $i$\textsuperscript{th} sample, and $n_{t}$ is the number of samples. The first domain, $\mathcal{D}_{1}$, is referred to as the source domain.
We assume that a deep learning model $h_{\theta_{1}}(.)$, where $\theta_1$ is the pre-trained parameters on the source data $\textbf{x}_{1,i}$ in $\mathcal{D}_1$. However, we discard the source data thereafter, typically because of privacy or memory concerns.
More precisely, we cannot access source data $\mathcal{D}_1$ for future new domains. 
Therefore, the future domains $\mathcal{D}_{t}$ with $t > 1$, are termed as the target domains. 
Within the CTDA framework, our objective is to improve the performance of the model $h_{\theta_{1}}(\textbf{x})$ during test time for a dynamically changing target domain in an online manner. This involves feeding the model unlabeled test data from the target domain sequentially.
To be more specific, at target domain $\mathcal{D}_{t}$, the unlabeled data $\textbf{x}_{t}$ is given to the model $h_{\theta_{t}}(\textbf{x}_{t})$, and the model needs to make the prediction and adapt itself accordingly ($\theta_{t} \to \theta_{t+1}$) for the next target domain, $\mathcal{D}_{t+1}$. 
In the process of test-time training, we ensure that the model is adapted to each current sample from the target domain. Following this adaptation, the updated model is utilized to predict the class label of the respective sample.

\subsection{Model Overview}
\begin{figure*}[!th]
\centering
\includegraphics[width=1.0\linewidth]{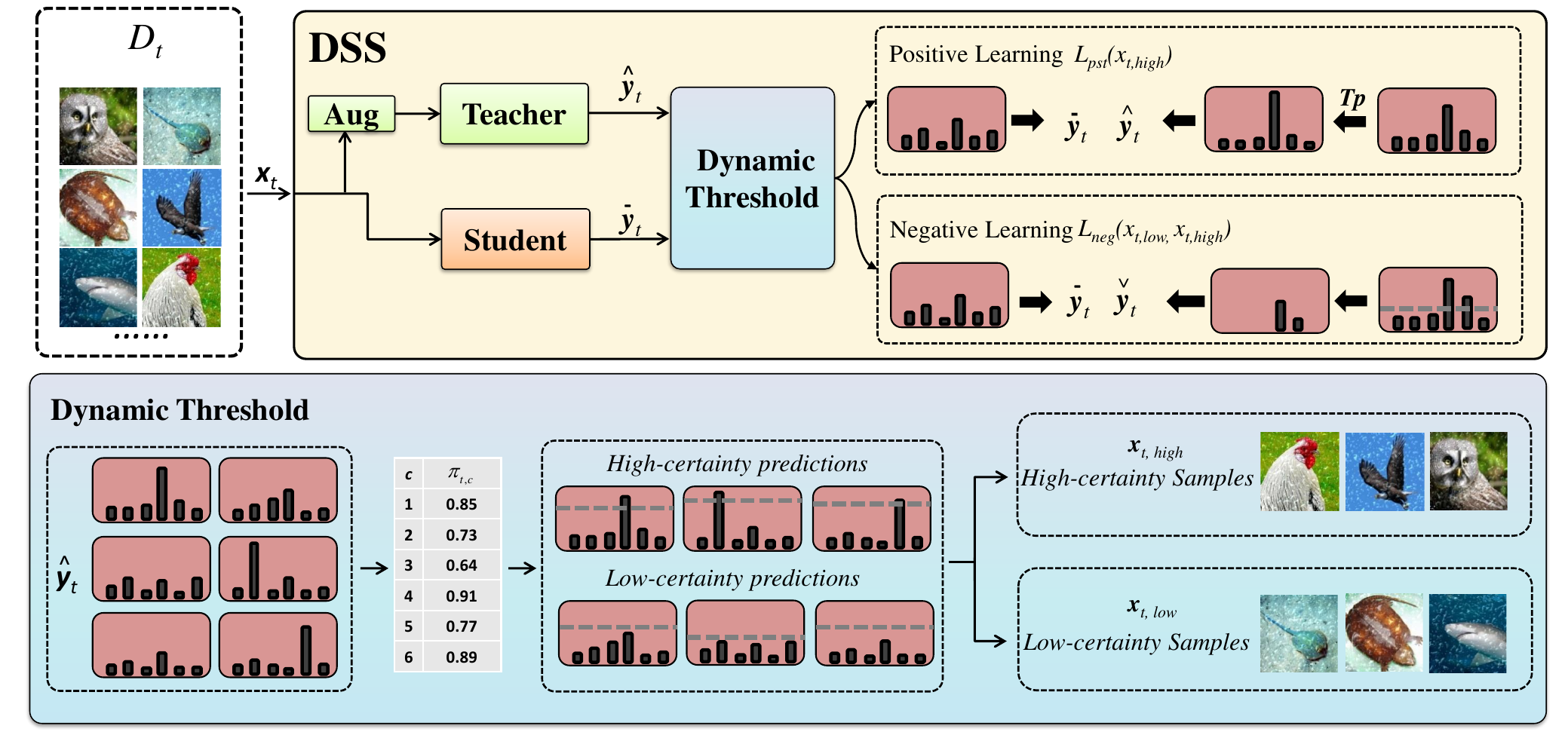}
\caption{{The proposed Dynamic Sample Selection (DSS) framework for continual test-time domain adaptation (CTDA). Unlabeled target domains are changing over time. At time step $t$, the model first makes the predictions for samples $\textbf{x}_{t}$ from $D_t$. Taking the input $\textbf{x}_{t}$ from domain $\mathcal{D}_t$, the teacher model computes the pseudo label $\hat{\textbf{y}}_{t}$ based on augmentation module while the student model gives the output $\Bar{\textbf{y}}_{t}$.  The dynamic threshold module then outputs a suitable threshold to split the samples $\textbf{x}_{t}$ into two groups, high- and low-quality groups. Positive learning is exclusively applied to the high-quality group of samples, ensuring that the model focuses its learning efforts on reliable and accurate data. Conversely, in negative learning, complementary labels are generated for classes that yield low prediction scores. This approach enables the model to learn from and address the challenging samples actively, contributing to improved performance and adaptability in evolving domains.}}

\label{fig:architecture_DSS}
\end{figure*}

This section provides a brief overview of our proposed architecture, Dynamic Sample Selection (DSS), which is illustrated in Figure~\ref{fig:architecture_DSS}. Similar to other methods in Continual Target Domain Adaptation (CTDA)\cite{wang2022continual}, we adopt the student and teacher setup from the mean teacher framework\cite{NIPS2017_68053af2} as the foundational structure for self-learning in our approach. Given the input $\textbf{x}_{t}$ from domain $\mathcal{D}t$, the teacher model computes the pseudo label $\hat{\textbf{y}}_{t}$ based on an augmentation module, while the student model produces the output $\Bar{\textbf{y}}_{t}$. The model is then trained using the loss function that enforces consistency between $\hat{\textbf{y}}_{t}$ and $\Bar{\textbf{y}}_{t}$, thus promoting alignment between the predicted labels. It is worth noting that $\hat{\textbf{y}}_{t}$ is typically used as the final prediction.

As indicated in Figure~\ref{fig:architecture_DSS}, we do not treat all samples in $\mathcal{D}_t$ equally for adaptation. In contrast, we assume that not all samples have enough quality for training during test time. 
To achieve this objective, we employ adaptive pseudo-labelling from recent advances in semi-supervised learning~\cite{zhang2021flexmatch,wang2022freematch}, a learning approach that involves continuous monitoring of the model's learning progress in accordance with its learning status~\cite{bengio2009curriculum}.
In other words, we utilize a dynamic threshold module to modulate the learning progress adaptively each time we encounter new domains. Therefore, we design a DSS framework on top of the mean teacher model to select samples for different training strategies. To be more specific, given input $\textbf{x}_{t}$ in $\mathcal{D}_t$, we leverage the student-teacher setup to generate the prediction (or pseudo) label $\hat{\textbf{y}}_{t}$. Using $\hat{\textbf{y}}_{t}$, we compute the adaptive class-wise threshold $\pi_{t,c}(\hat{\textbf{y}}_{t})$ for class $c$. Based on $\pi_{t,c}$, we group high- and low-quality samples, $\textbf{x}_{t,high}$ and $\textbf{x}_{t,low}$. 
Then, both $\textbf{x}_{t,high}$ and $\textbf{x}_{t,low}$ are used for negative learning. Meanwhile, the model is trained with the positive learning loss with high-quality samples $\textbf{x}_{t, high}$ alone. 

\subsection{Dynamic Thresholding}
In CTDA, the learning progress of the model $h_{\theta}$ varies over time since different target domains have significant domain gaps. When adopting a model to new test samples from different domains, pseudo labels inevitably suffers from label noise due to the presence of domain shifts. To minimize noise in test-time, We adopt a threshold mechanism to adjust the confidence threshold dynamically.

As shown in Figure~\ref{fig:architecture_DSS}, the samples $\textbf{x}_{t}$ from $\mathcal{D}_t$ are first given to the teacher model to compute the initial predictions $\hat{\textbf{y}}_{t}$. 
Subsequently, taking inspiration from the threshold strategy~\cite{zhang2021flexmatch, wang2022freematch} and in order to track the learning progress, we implement a dynamic approach that involves detecting low-confidence predictions in samples through the use of a dynamic threshold.
The common practice is to use a fixed confidence threshold to remove suspected noisy labels. However, model confidence on out-of-distribution samples could differ vastly for each domain, which may hurt the adaptation performance under a fixed threshold. Thus, we propose the adaptive threshold function to output a suitable selection threshold $\pi_{t}$:  
\begin{equation}
\pi_{t, j} = \lambda\pi_{t, j-1} + (1-\lambda) \cdot \frac{1}{N} \sum^{N}_{i=1} {\max}~ \hat{\textbf{y}}_{t,j,i}
\end{equation}
where $j$ denotes the order of the batch. For simplicity, we drop the most $j$ in the following context. $\pi_{t}$ denotes the adaptive threshold value for the samples $\textbf{x}_{t}$. $\hat{\textbf{y}}_{t,i}$ is the predicted confidence vector for $i$\textsuperscript{th} sample. $\max \hat{\textbf{y}_{t,i}}$ indicates the highest confidence value in $\hat{\textbf{y}_{t,i}}$. The number of samples from each batch is denoted as $N$, and $\lambda$ represents the exponential moving average factor. 


An appropriate initial threshold could help the algorithm to converge more swiftly to an effective threshold in an online manner. It is important to note that DSS algorithm operates in an online fashion during the inference phase. Starting with a threshold value that is considerably distant from the current confidence level may result in a significant time delay before reaching a satisfactory threshold. Therefore, we adopt the following approach to initialize the threshold at the start of each domain: we take the average of the previous model confidence at the beginning, denoted as $\pi_{t,0} = (\pi_{t-1, final} + \frac{1}{C})/2$ where $\pi_{t-1, final}$ is the computed threshold from the previous domain and $C$ represents the number of classes. It should be mentioned that for the first domain, we set $\pi_{0,0} = \frac{1}{C}$. By incorporating the average of the previous domain's threshold, we aim to achieve an accurate estimation of the initial threshold by leveraging the knowledge acquired from the previous domains. Unlike other existing methods in semi-supervised learning\cite{berthelot2019mixmatch}, we use flexible threshold initialization in CTDA to provide a more appropriate initial value when facing a new domain instead of using a fixed average constant.

To reflect the discrepancy between the adaptation status among different classes, we further give the threshold in a class-wise manner to ensure that the class with average low confidence (low-certainty class) is not ignored. The class-wise manner guarantees sufficient samples for training as there are many low-certainty classes. The aim is to prevent the model from being biased toward the majority classes. We first compute the rescale ratio $\tau_{t,c}$ for class $c$ as follows: 
\begin{equation}
\tau_{t,c} = \frac{\delta_{t,c}}{{\max} \bm{\delta}_{t}}
\end{equation}
where $\bm{\delta}_{t}$ is the vector where one element is the average confidence for one class. $\delta_{t,c}$ is the average confidence for class $c$ of samples in $D_t$. The formulation $\delta_{t,c}$ is given as follows:
\begin{equation}
\delta_{t,c} = \frac{1}{N} \sum^{N}_{i=1} {\max}~ \hat{\textbf{y}}_{t,i} \cdot \mathbbm{1}(\arg\max\hat{\textbf{y}}_{t,i}={c}) 
\end{equation}
where ${\max}~ \hat{y}_{t,i}$ is the maximum prediction probability in the vector $\hat{y}_{t,i}$. Hence, the improved version of the adaptive threshold in a class-wise manner is calculated as:
\begin{equation}\label{eq:adpt_threshold}
\pi_{t, c} = \pi_{t} \cdot \tau_{t, c}
\end{equation}

For $\mathcal{D}_t$, after achieving $\pi_{t, c}$ for samples, we divide $\textbf{x}_{t}$ into the high- and low-quality groups, $\textbf{x}_{t,high}$ and $\textbf{x}_{t,low}$. $\textbf{x}_{t,high}$ are samples where their predicted confidence is above the threshold $\pi_{t, c}$, and vice versa.

In the following algorithmic steps, the different training strategies will be used for $\textbf{x}_{t, high}$ and $\textbf{x}_{t, low}$. Overall, the self-adaptive threshold accurately identifies correctly- and wrongly-predicted samples as high- and low-certainty groups. Grouped samples will facilitate positive and negative learning, introduced in the following subsection.


\subsection{Joint Positive and Negative Learning}
Training the model in CTDA without access to data labels is challenging. To be more specific, assigning the computed pseudo label to the unlabeled data runs the risk of letting the model learn incorrect information as the pseudo label is noisy, which could lead to lower overall accuracy. To address this issue, as shown in Figure \ref{fig:architecture_DSS} we first group samples into high- and low-quality groups using dynamic threhold. Then, we train two groups of samples in two different ways (See ``DSS: Positive and Negative Learning" of Figure \ref{fig:architecture_DSS}). 

More specifically, we restrict positive learning to the high-quality group $\textbf{x}_{t,high}$, as these samples are expected to have a higher probability of being accurately classified by the network. Conversely, negative learning~\cite{kim2021joint, kim2019nlnl} is applied to both groups (\ie, $\textbf{x}_{t,high}$ and $\textbf{x}_{t,low}$) to mitigate the risk of providing inaccurate information to the model by generating the complementary labels for the training.

To train our proposed DSS, we employ a cross-entropy loss function. We first have the pseudo-label generated by the teacher model $\hat{\textbf{y}}_{k, high}$, and then we apply temperature scaling to minimize the prediction entropy by sharpening the prediction distribution~\cite{berthelot2019mixmatch}.

\begin{equation}
\operatorname{Sharpen}(\hat{\textbf{y}}_{k, high}, Tp):=\hat{\textbf{y}}_{k, high}^{{Tp}} / \sum_{k=1}^C \hat{\textbf{y}}_{k, high}^{{Tp}}
\end{equation}
where $Tp$ is the temperature hyper-parameter. 
Most existing methods construct pseudo labels in one-hot way~\cite{sohn2020fixmatch, zhang2021flexmatch}. However, in CTDA, we mildly give pseudo labels by applying the sharpening. Specifically, we propose to minimize entropy by constructing sharpened pseudo labels with $Tp$ for selected high-certainty samples. It somewhat avoids overfitting by preserving prediction structures from the teacher model.

The loss function for positive learning is formulated as the cross entropy loss between the student and teacher predictions:
\vspace{-5mm}

\begin{equation} \label{eq:pst_loss}
\mathcal{L}_{pst}(\Bar{\textbf{y}}_{k,high}, \hat{\textbf{y}}_{k,high})
=-\sum_{k=1}^C \hat{\textbf{y}}_{k,high} \log \Bar{\textbf{y}}_{k,high}
\end{equation}

where is $\Bar{\textbf{y}}_{k,high}$ the prediction from the student model. This loss function will enforce alignment between the student prediction and the more accurate teacher prediction after sharpening. To adequately utilize $\hat{\textbf{y}}$, the complementary label is also created based on low class confidences. The complementary label represents "the input does not belong to particular classes." This provides more reliable information regarding the classes that are most likely not to be the true label. Following this, we present the negative learning loss as follows:
%
\begin{equation} \label{eq:neg_loss}
\mathcal{L}_{neg}(\Bar{\textbf{y}}_{k}, \hat{\textbf{y}}_{k})
=-\sum_{k=1}^C \check{\textbf{y}}_{k} \log \left(1-\Bar{\textbf{y}}_k\right)
\end{equation}
where $\check{\textbf{y}}_{k} = \mathbbm{1}(\hat{\textbf{y}}_k<\bm{\alpha})$ is the complementary label based on the prediction from the teacher model, $\hat{\textbf{y}}_{k}$. Applying the complementary label $\check{\textbf{y}}$ in the negative learning pushes the prediction away from those classes with low confidence. $\bm{\alpha}$ is the threshold vector whose element value is set to $0.05$. Notably, we keep the teacher model fixed and back-propagate the student model only. The reason is that we aim to ensure a stable student-teacher setup.
Finally, having the Equations \ref{eq:pst_loss} and \ref{eq:neg_loss} at hand, we build the total loss function as follows:
\begin{equation}\label{eq:total_loss}
\mathcal{L}_{total}=\mathcal{L}_{neg}+\mathcal{L}_{pst} 
\end{equation}

Moreover, the teacher model with $\hat{\theta}$ is updated by the moving average of the parameters of the student model, $\Bar{\theta}$. Hence, we have
\begin{equation}\label{eq:mv_average}
\hat{\theta}_{t+1}=\beta \hat{\theta}_t + (1-\beta) \Bar{\theta}_{t+1}
\end{equation}
where $\beta$ is the coefficient. We set $\theta$ to 0.999 which is the same as that in \cite{wang2022continual}. The overall training process is described in Algorithm~\ref{alg:method}.


\begin{algorithm}[!t]
\caption{The proposed DSS for CTDA}\label{euclid}
\begin{algorithmic}[1]
\Statex \textbf{Initialization:} 
A source model $h_{\theta_{1}}$ pre-trained on source domain $\mathcal{D}_{1}$, teacher model $h^{t}_{\theta_{t}}$ and student model $h^{s}_{\theta_{t}}$ initialized from $h_{\theta_{1}}$.

\Statex \textbf{Input:} Unlabeled test data $\textbf{x}_{t}$ for target domain $\mathcal{D}_{t}$, $t>1$,
\For{$t=2$ to $T$}

\State \hspace*{-1mm}Augment $\textbf{x}_{t}$  and get pseudo-labels $\hat{\textbf{y}}_{t}$;
\State \hspace*{-1mm}Generate class-wise threshold $\pi_{t, c}$ via Eq. (\ref{eq:adpt_threshold});
\State \hspace*{-1mm}Divide $\textbf{x}_{t}$ into  $\textbf{x}_{t,low}$ and $\textbf{x}_{t,high}$ by  $\pi_{t,c}$;
\State \hspace*{-1mm}Compute $\mathcal{L}_{pst}$ via Eq. (\ref{eq:pst_loss});
\State \hspace*{-1mm}Compute $\mathcal{L}_{neg}$ via Eq. (\ref{eq:neg_loss});
\State \hspace*{-1mm}Update student model via $\mathcal{L}_{total}$ in Eq. (\ref{eq:total_loss});
\State \hspace*{-1mm}Update teacher model via Eq. (\ref{eq:mv_average});
\Statex \textbf{Output:} Prediction $\hat{\textbf{y}}_{t}$, updated student and teacher models $h^{s}_{\theta_{t}}$ and $h^{t}_{\theta_{t}}$, selection threshold $\pi_{t,c}$.

\EndFor
\end{algorithmic}
\label{alg:method}
\end{algorithm}
\vspace{-2mm}

\section{Experiments}
Extensive experiments are conducted in this section to demonstrate the effectiveness of our DSS approach. We evaluate our method on three 2D image continual test-time adaptation benchmark tasks, CIFAR-10-C \cite{hendrycks2019benchmarking}, CIFAR-100-C \cite{hendrycks2019benchmarking}, and ImageNet-C \cite{hendrycks2019benchmarking}, designed to assess the robustness of machine learning models to corruptions and perturbations in the input data. In addition, we test our proposed method on a 3D point cloud dataset ScanObjectNN-C derived from ScanObjectNN~\cite{uy-scanobjectnn-iccv19}. 

\subsection{Datasets}
\noindent\textbf{CIFAR10-C.} CIFAR10-C is an extension of the CIFAR-10 dataset \cite{krizhevsky2009learning}, which consists of $32 \times 32$ color images from 10 classes. CIFAR10-C includes 15 different corruptions, each at five levels of severity. The corruptions are applied to test images of CIFAR-10, resulting in a total of 10,000 images.

\noindent\textbf{CIFAR100-C.} CIFAR100-C is an extension of the CIFAR-100 dataset \cite{krizhevsky2009learning}, which consists of $32 \times 32$ color images in 100 classes. CIFAR100-C includes 15 different corruptions, each at five levels of severity. The corruptions are applied to the test images of CIFAR-100, resulting in a total of 10,000 images.

\noindent\textbf{ImageNet-C.} ImageNet-C is an extension of the ImageNet dataset~\cite
{deng2009imagenet}, which contains over 14 million images in more than 20,000 categories. ImageNet-C includes 15 different corruptions, with 5 severity levels. The corruptions are applied to the validation images of ImageNet.

\noindent\textbf{ScanObjectNN-C.} ScanObjectNN~\cite{uy-scanobjectnn-iccv19} is a point cloud classification dataset that is collected from the real world. It contains 15 classes, with 2309 samples in the train set and 581 in the test set. To build ScanObjectNN-C, we employ the setting proposed by~\cite{sun2022benchmarking}, to generate 15 corruptions in the test set of ScanObjectNN for our experiments. 

\subsection{Implementation} 
We strictly follow the setting of CTDA that no source data is accessed \cite{wang2022continual}. All models are evaluated based on the largest corruption severity level of five for all datasets in an online fashion. Model predictions are first generated before adapting to the current test stream.

Following \cite{wang2022continual}, we adopt standard pre-trained WideResNet \cite{BMVC2016_87}, ResNeXt-29 \cite{yin2019fourier} and ResNet-50 \cite{croce2020robustbench} on CIFAR10-C, CIFAR100-C, and ImageNet-C as the source models. The sharpening factor $Tp$ used in temperature scaling is set to $0.6$, and augmentation module from \cite{wang2022continual} is used to generate the augmentation-weighted pseudo-label. For 3D experiments, we use DGCNN \cite{wang2019dynamic} pre-trained on the clean set as the backbone. 

\begin{table*}[th]
\centering
\resizebox{1.0\textwidth}{!}{
\begin{tabular}{l|c|c|c|c|c|c|c|c|c|c|c|c|c|c|c|c}
\hline
\textbf{Method} & \rotatebox{0}{\textbf{Gaussian}}& \rotatebox{0}{\textbf{shot}} &\rotatebox{0}{\textbf{impulse}} & \rotatebox{0}{\textbf{defocus}} &\rotatebox{0}{\textbf{glass}} & \rotatebox{0}{\textbf{motion}} &\rotatebox{0}{\textbf{zoom}} &\rotatebox{0}{\textbf{snow}} &\rotatebox{0}{\textbf{frost}} & \rotatebox{0}{\textbf{fog}} & \rotatebox{0}{\textbf{brightness}} & \rotatebox{0}{\textbf{contrast}} &\rotatebox{0}{\textbf{elastic\_trans}}&\rotatebox{0}{\textbf{pixelate}} & \rotatebox{0}{\textbf{jpeg}} & \textbf{Mean} \\ \hline
\textbf{Source}        &72.3 & 65.7 &72.9 &46.9 &54.3 &34.8 &42.0 &25.1 &41.3 &26.0 &9.3 &46.7 &26.6 &58.5 &30.3 &43.5 \\ 
\textbf{BN Adapt\cite{li2016revisiting}}          &28.1 &26.1 &36.3 &12.8 &35.3 &14.2 & 12.1 &17.3 &17.4 &15.3 &8.4 &12.6 &23.8 & 19.7 &27.3 &20.4\\ 
\textbf{TENT-cont} \cite{wang2021tent}     &24.8 &\textbf{20.6} &28.6 &14.4 &31.1 &16.5 &14.1 &19.1 &18.6 &18.6 &12.2 &20.3 &25.7 &20.8 &24.9 &20.7\\ 
\textbf{AdaContrast\cite{chen2022contrastive}}   &29.1 &22.5 &30.0 &14.0 &32.7 &14.1 &12.0 &16.6 &14.9 &14.4 &8.1 &\textbf{10.0} &21.9 &17.7 &20.0 &18.5 \\
\textbf{Conjugate PL \cite{goyal2022test}} &27.6 &25.7 &35.9 &12.7 &34.8 &14.1 &12.0 &17.1 &17.3 &15.0 &8.4 &12.3 &23.5 &19.3 &26.7 &20.2 \\
\textbf{CoTTA} \cite{wang2022continual} &24.3 &21.3 &26.6 &\textbf{11.6} &27.6 &12.2 &\textbf{10.3} &14.8 &14.1 &\textbf{12.4} &\textbf{7.5} &10.6 &18.3 & 13.4 &17.3 &16.2 \\  \hline
\textbf{DSS (Ours)}  &\textbf{24.1} &21.3 &\textbf{25.4} &11.7&\textbf{26.9} &\textbf{12.2} &10.5 &\textbf{14.5} &\textbf{14.1} &12.5 &7.8 &10.8 &\textbf{18.0} & \textbf{13.1} &\textbf{17.3} &\textbf{16.0} \\
\hline

\end{tabular}}
\caption{Classification error rate (\%) on CIFAR10-C. The best numbers are in bold.
}\label{tab:cifar10-c}
\end{table*}

\begin{table*}[th]
\centering
\resizebox{1.0\textwidth}{!}{
\begin{tabular}{l|c|c|c|c|c|c|c|c|c|c|c|c|c|c|c|c}
\hline
\textbf{Method} & \rotatebox{0}{\textbf{Gaussian}}& \rotatebox{0}{\textbf{shot}} &\rotatebox{0}{\textbf{impulse}} & \rotatebox{0}{\textbf{defocus}} &\rotatebox{0}{\textbf{glass}} & \rotatebox{0}{\textbf{motion}} &\rotatebox{0}{\textbf{zoom}} &\rotatebox{0}{\textbf{snow}} &\rotatebox{0}{\textbf{frost}} & \rotatebox{0}{\textbf{fog}} & \rotatebox{0}{\textbf{brightness}} & \rotatebox{0}{\textbf{contrast}} &\rotatebox{0}{\textbf{elastic\_trans}}&\rotatebox{0}{\textbf{pixelate}} & \rotatebox{0}{\textbf{jpeg}} & \textbf{Mean}\\\hline
\textbf{Source}&73.0&68.0&39.4&29.3&54.1&30.8&28.8&39.5&45.8&50.3&29.5&55.1&37.2&74.7&41.2&46.4\\
\textbf{BN Adapt \cite{li2016revisiting}}  &42.1&40.7& 42.7& 27.6& 41.9& 29.7& 27.9& 34.9& 35.0& 41.5& 26.5& 30.3& 35.7& 32.9& 41.2& 35.4\\
\textbf{TENT-cont} \cite{wang2021tent} &\textbf{37.2}& \textbf{35.8}& 41.7& 37.9& 51.2& 48.3& 48.5& 58.4& 63.7& 71.1& 70.4& 82.3& 88.0& 88.5& 90.4&60.9\\
\textbf{AdaContrast \cite{chen2022contrastive}} &42.3& 36.8& 38.6& 27.7& 40.1& 29.1& 27.5& 32.9& 30.7& 38.2& 25.9& 28.3& 33.9& 33.3& 36.2& 33.4\\
\textbf{ Conjugate PL \cite{goyal2022test}} & 39.2& 37.1& 36.9& \textbf{26.3}& 39.4& 28.4& 26.4& 32.9& 33.2& 38.2& 25.7& 29.3& 34.0& 30.3& 39.0& 33.1\\
\textbf{CoTTA} \cite{wang2022continual} &40.1& 37.7& 39.7& 26.9& 38.0& 27.9& 26.4& 32.8& 31.8& 40.3& 24.7& 26.9& 32.5& 28.3& 33.5& 32.5\\\hline
\textbf{DSS (Ours)}&39.7&36.0&\textbf{37.2}&\textbf{26.3}&\textbf{35.6}&\textbf{27.5}&\textbf{25.1}&\textbf{31.4}&\textbf{30.0}&\textbf{37.8}&\textbf{24.2}&\textbf{26.0}&\textbf{30.0}&\textbf{26.3}&\textbf{31.1}&\textbf{30.9} \\

\hline
\end{tabular}}
\caption{Classification error rate (\%) on CIFAR100-C. The best numbers are in bold.
}\label{tab:cifar100-c}
\end{table*}


\begin{table*}[!t]
\centering
\resizebox{1.0\textwidth}{!}{
\begin{tabular}{l|c|c|c|c|c|c|c|c|c|c|c|c|c|c|c|c}
\hline 
\textbf{Method} & \rotatebox{0}{\textbf{Gaussian}}& \rotatebox{0}{\textbf{shot}} &\rotatebox{0}{\textbf{impulse}} & \rotatebox{0}{\textbf{defocus}} &\rotatebox{0}{\textbf{glass}} & \rotatebox{0}{\textbf{motion}} &\rotatebox{0}{\textbf{zoom}} &\rotatebox{0}{\textbf{snow}} &\rotatebox{0}{\textbf{frost}} & \rotatebox{0}{\textbf{fog}} & \rotatebox{0}{\textbf{brightness}} & \rotatebox{0}{\textbf{contrast}} &\rotatebox{0}{\textbf{elastic\_trans}}&\rotatebox{0}{\textbf{pixelate}} & \rotatebox{0}{\textbf{jpeg}} & \textbf{Mean}\\\hline
\textbf{Source} &95.3&94.6 &95.3&84.9&91.1&86.8&77.2&84.4&80.0&77.3&44.4&95.6&85.2&76.9&66.7&77.2\\
\textbf{BN Adapt} \cite{li2016revisiting} &87.6&87.4  &87.8&87.7&88.0&78.2&64.5&67.6&70.6&54.9&\textbf{36.4}&89.3&58.0&56.4&66.6& 66.2\\
\textbf{TENT-cont} \cite{wang2021tent} &85.7&80.0&78.3&\textbf{82.2}&\textbf{79.2}&70.9&\textbf{59.1}&65.6&66.4&55.4&40.6&80.3&55.5&53.5&59.0&67.4 \\
\textbf{ Conjugate PL \cite{goyal2022test}} & 85.2& \textbf{79.6}& \textbf{77.1}& 82.4& 79.8& \textbf{70.8}& 59.3& 65.2& 66.1& 54.8& 39.8& 79.5& 55.1& 52.7& 58.7& 67.1 \\
\textbf{CoTTA} \cite{wang2022continual} & 87.5&86.0&84.4&85.1&84.4&73.9&61.5&63.6&64.2&51.9&38.6&74.8&51.1&45.1&50.2&66.8  \\ \hline
\textbf{DSS (Ours)}
&\textbf{84.6}&80.4&78.7&83.9&79.8&74.9&62.9&\textbf{62.8}&\textbf{62.9}&\textbf{49.7}&37.4&\textbf{71.0}&\textbf{49.5}&\textbf{42.9}&\textbf{48.2}&\textbf{64.6}\\ \hline

\end{tabular}}
\caption{Classification error rate (\%) on ImageNet-C. The best numbers are in bold.
} \label{tab:imagenet-c}
\end{table*}

\subsection{Main Results}
\noindent\textbf{2D Results.}
We first examine the adaptation performance on 2D tasks, and the experimental results are summarized in Table~\ref{tab:cifar10-c}, \ref{tab:cifar100-c}, and ~\ref{tab:imagenet-c}. 
Directly testing the source model on target domains in sequence yields high average errors of $43.5\%$, $46.4\%$, and $77.2\%$ on CIFAR10-C, CIFAR100-C, and Imagenet-C respectively. Applying BN Stats Adapt \cite{li2016revisiting} to update the batch normalization statistics from the current test stream, the average error across all target domains is significantly reduced on all datasets. The TENT-based method \cite{wang2021tent} also helps the model to adapt to the target domain in sequence, but it may suffer severe error accumulation in the long term.As shown in Table~\ref{tab:cifar100-c}, the TENT-based method \cite{wang2021tent} yields a substantially higher error rate of $60.9\%$ in the long run on CIFAR100-C. Similarly, Conjugate PL~\cite{goyal2022test} perform well in early adaptation to multiple initial domains, but soon experiences a gradual increase in error rate over time. Based on more accurate weighted augmentation-averaged predictions, CoTTA \cite{wang2022continual} has the lowest average error rate in comparison with other adaptation methods. 

By leveraging joint positive and negative learning with an adaptive threshold, our proposed method DSS constantly outperforms its baseline CoTTA in all datasets. This indicates that DSS helps the model better adapt to continual target domains with less suffering from error accumulation generated from noisy pseudo labels. We notice that this phenomenon becomes more evident in more difficult datasets, where the model has low certainty on proceeding target domains. Compared with CoTTA, DSS successfully reduces the average error from $32.5\%$ to $30.9\%$ and  $66.8\%$ to $64.6\%$ on CIFAR100-C and ImageNet-C respectively, which demonstrates the advantage of our DSS module. 

To further investigate the effectiveness of DSS over the baseline, we also evaluate its adaptation performance over ten different sequences on ImageNet-C (See Table~\ref{tab:imagenet-c-sequence}). There is a $1.4\%$ decrease on average over $10$ diverse sequences in comparison with CoTTA, indicating that our method is more robust to the order of the target domain sequence.

\begin{table*}[t]
\centering
\resizebox{0.85\textwidth}{!}{
\begin{tabular}{l|c|c|c|c|c|c}
\hline 
\textbf{Avg. Error (\%)} & \rotatebox{0}{\textbf{Source}}& \rotatebox{0}{\textbf{BN Adapt}} \cite{li2016revisiting} &\rotatebox{0}{\textbf{Test Aug}} & \rotatebox{0}{\textbf{TENT}} \cite{wang2021tent} &\rotatebox{0}{\textbf{CoTTA}} \cite{wang2022continual} & \rotatebox{0}{\textbf{DSS (Ours)}} \\\hline
\textbf{ImageNet-C}    &82.4 &72.1 &71.4&66.5&63.0 (±$1.8$) &\textbf{61.6} (±$0.3$)\\\hline
\end{tabular}}
\caption{Classification error rate (\%) on ImageNet-C over 10 sequences. The best numbers are in bold.} \label{tab:imagenet-c-sequence}
\end{table*}

\begin{table*}[!t]
\centering
\resizebox{1.0\textwidth}{!}{
\begin{tabular}{l|c|c|c|c|c|c|c|c|c|c|c|c|c|c|c|c}
\hline 
\textbf{Method} & \rotatebox{0}{\textbf{uniform}}& \rotatebox{0}{\textbf{gaussian}} &\rotatebox{0}{\textbf{background}} & \rotatebox{0}{\textbf{impulse}} &\rotatebox{0}{\textbf{upsampling}} & \rotatebox{0}{\textbf{rbf}} &\rotatebox{0}{\textbf{rbf-inv}} &\rotatebox{0}{\textbf{den-dec}} &\rotatebox{0}{\textbf{dens-inc}} & \rotatebox{0}{\textbf{shear}} & \rotatebox{0}{\textbf{rot}} & \rotatebox{0}{\textbf{cut}} &\rotatebox{0}{\textbf{distort}}&\rotatebox{0}{\textbf{oclsion}} & \rotatebox{0}{\textbf{lidar}} & \textbf{Mean}\\\hline
\textbf{Source}&58.7&49.6&55.4&43.5&53.9&37.5&33.2&24.1&19.6&32.9&38.4&24.8&34.8&91.7&92.6&46.0 \\
\textbf{TENT-cont} \cite{wang2021tent}   &53.2&41.3&59.9&46.5&49.7&37.7&36.0&31.0&26.3&37.5&40.3&34.8&39.1&90.5&90.9&47.6 \\
\textbf{BN \cite{li2016revisiting}}    &\textbf{51.3}&\textbf{41.1}&\textbf{53.0}&48.7&\textbf{49.6}&34.8&31.7&29.4&22.2&32.9&37.2&32.4&34.8&90.2&90.4&45.3\\
\textbf{CoTTA} \cite{wang2022continual}  &56.6&47.2&57.0&39.2&\textbf{49.6}&33.2&\textbf{31.0}&\textbf{21.2}&16.2&29.8&34.6&\textbf{21.3}&30.3&90.9&91.4&43.3\\\hline

\textbf{DSS (Ours)} &56.6&46.8&57.1&\textbf{39.1}&\textbf{49.6}&\textbf{32.9}&\textbf{31.0}&\textbf{21.2}&\textbf{16.0}&\textbf{29.8}&\textbf{34.1}&21.9&\textbf{30.1}& \textbf{89.8}&\textbf{90.0}&\textbf{43.1}\\ \hline 

\end{tabular}}
\caption{Classification error rate (\%) on ScanObjectNN-C. DGCNN \cite{wang2019dynamic} is adopted as the backbone. The best numbers are in bold.
} \label{tab:ScanObjectNN-C-DGCNN}
\end{table*}

\noindent\textbf{3D Results.}
While our method is designed to address the CTDA task for 2D image recognition, we also adapt and evaluate our method for the case of 3D point cloud recognition \cite{qi2017pointnet, wang2019dynamic, hong2023pointcam}. To this end, we evaluate our proposed methods on ScanObject-C. As shown in Table~\ref{tab:ScanObjectNN-C-DGCNN}, DSS is again able to reach the lowest error rate of $43.1\%$ using DGCNN \cite{wang2019dynamic} as the backbone. The CoTTA method, on the other hand, performs $0.2\%$ worse on average due to the rapid accumulation of errors during the adaptation process.

\subsection{Ablation Study}
\begin{table}[t]
\centering
\resizebox{0.48\textwidth}{!}{
\begin{tabular}{l|c|c|c}
\hline 
\textbf{Avg. Error (\%)} & \rotatebox{0}{\textbf{CIFAR10-C}}&  \rotatebox{0}{\textbf{CIFAR100-C}} & \rotatebox{0}{\textbf{Imagenet-C}} \\\hline

\textbf{CoTTA} \cite{wang2022continual} &16.2 &32.5 & 66.8 \\ \hline
\textbf{DSS (\textit{w/} DT)}         &16.1&31.8&66.0 \\
\textbf{DSS (\textit{w/} DT\&PL)}     &16.1&31.0&64.8 \\
\textbf{DSS (\textit{w/} DT\&PL\&NL)} &\textbf{16.0}&\textbf{30.9}&\textbf{64.6} \\\hline
\end{tabular}}
\caption{Component analysis. DT, PL, and NL are dynamic thresholding, positive and negative learning. DSS (\textit{w/} DT) indicates solely using the high-quality samples and training the model with $\mathcal{L}_{cst}$, as CoTTA does. DSS (\textit{w/} DT\&PL) and (\textit{w/} DT\&PL\&NL) are trained with $\mathcal{L}_{pst}$ and $\mathcal{L}_{neg}+ \mathcal{L}_{pst}$, respectively.} \label{tab:componet_analysis}
\end{table} 

\noindent\textbf{Component analysis.} Our method selectively matches the standard student distribution, in contrast to CoTTA which attempts to align prediction probability for all test data. Aligning distributions for all data leads to error accumulation, as demonstrated in Table~\ref{tab:componet_analysis} where CoTTA naively matching distributions without considering the noisy condition of pseudo labels consistently generates larger errors over the continual domains. In comparison, we demonstrate that using our adaptive threshold (DSS \textit{w}/DT) can yield additional improvements of $0.7\%$ and $0.8\%$ on CIFAR100-C and Imagenet-C respectively. Furthermore, the positive learning (DSS \textit{w}/DT\&PL) leads to $0.8\%$ and $1.2\%$ improvements respectively over DSS \textit{w}/DT on CIFAR100-C and Imagenet-C. By applying negative learning (DSS \textit{w}/DT\&PL\&NL) to utilize more unlabelled low-confidence samples, we achieve further improvements of approximately $0.1\%$ and $0.2\%$. Apart from these, we observe that the overall improvement in CIFAR10-C is marginal compared with other cases, and this may due to the fact that the model is already very confident on those related target domains with average confidence greater than $90\%$. In that case, the source model is able to produce low-entropy predictions even without online adaptation, and our approach has a limited effect to overcome the bias that lies in the high-confidence predictions.

\begin{figure}[!t]\centering
\includegraphics[width=1.0\linewidth,trim=0cm 0cm 0cm 0cm, clip]{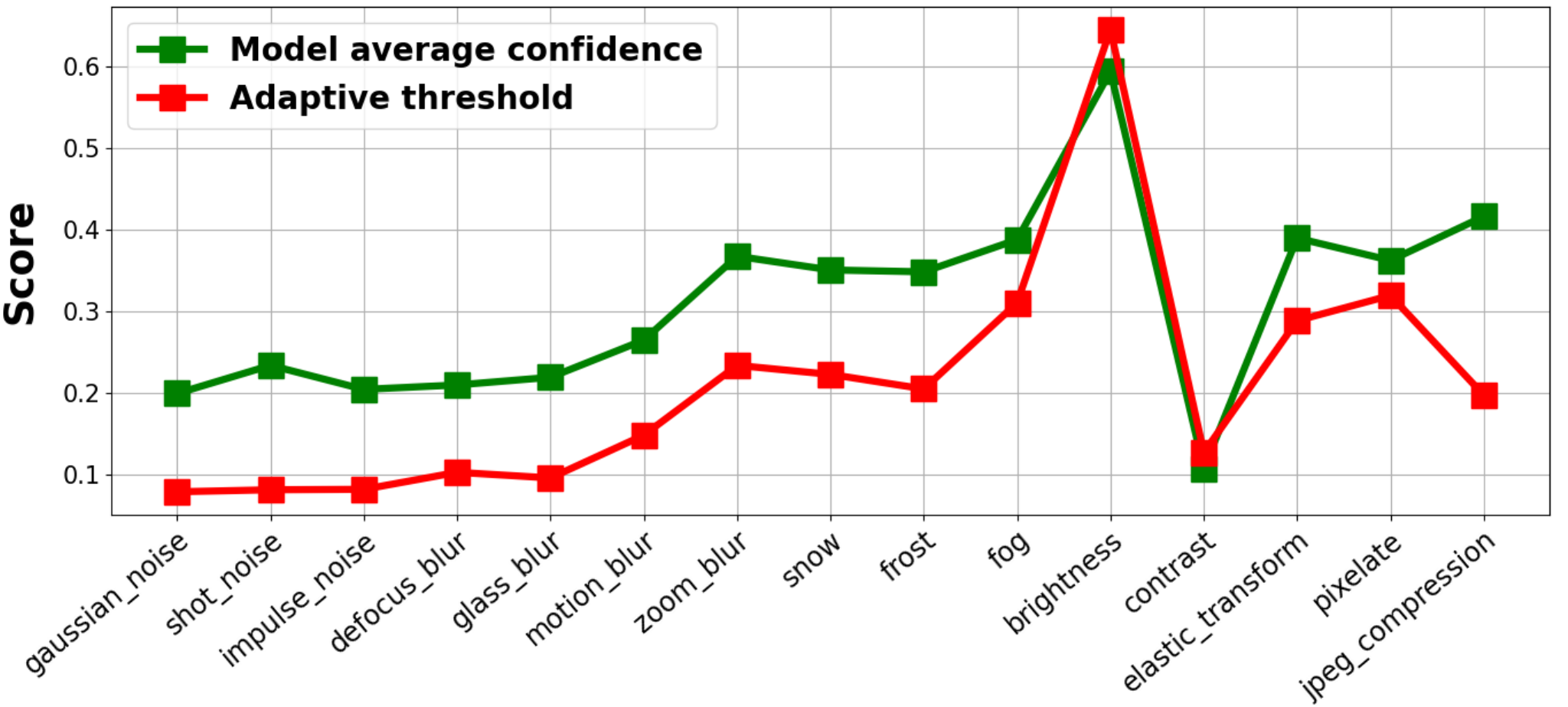}
\caption{ The average prediction confidence (green line) and adaptive threshold (red line) on ImageNet-C across domains.}
\label{fig:model_confidence}
\vspace{-6pt}
\end{figure}

\noindent\textbf{Threshold.}
Here, we validate the effectiveness of the dynamic threshold via comparisons with the simple fixed threshold approach across all domains. The results, as presented in Table~\ref{tab:Fixed Threshold}, first show that using a fixed threshold with $\pi = 0.2$ may help to reduce the average error rate by $0.1\%$ on CIFAR100-C. However, using a fixed threshold still causes incorrect predictions to be utilized for self-learning, resulting in a higher error rate of $0.6\%$ compared with the dynamic threshold. Note that the averaged model confidence on out-of-distribution data clearly varies across all target domains (See the green line of  Figure~\ref{fig:model_confidence}), and an inappropriate fixed threshold could limit the generalization performance. To illustrate this point, in Table~\ref{tab:Fixed Threshold}, we also show that fixed $\pi = 0.2$ actually increases the error rate to $67.8\%$ from $66.8\%$ on ImageNet-C, whereas our dynamic threshold still consistently decrease the error rate. This is mainly due to the fact that a fixed threshold may filter out excessive samples, leaving inadequate data for model adapation. Thus, we argue that our adaptive threshold approach strikes a better balance between maximizing target adaptation performance and minimizing error accumulation. 

Moreover, applying positive learning with a fixed threshold on ImageNet-C instead resulted in an increased average error rate of $77.4\%$ from $66.8\%$, indicating that naive entropy minimization by positive learning would lead to an increase in error accumulation rate. As such, the positive learning component can only be effectively applied with an appropriate threshold that filters out most incorrect pseudo labels. Our adaptive threshold module, however, can always provide a reliable value to filter suspected noisy labels.

\noindent\textbf{Sharpening.}
Sharpening is employed to decrease the entropy of teacher predictions to assist with model generalization. Here, we demonstrate that in Table~\ref{tab:temperture} when $Tp = 0.6$, it has the lowest average error rate over all domains. Nevertheless, the average error rate increases when the $Tp$ value falls below or exceeds $0.6$. The reasoning behind this phenomenon may be attributed to two different factors. Firstly, as $Tp$ approaches $1$, the model's ability to gain knowledge from predictions with higher entropy is limited, resulting in lower accuracy. On the other hand, decreasing $Tp$ below $0.6$ may push the models to output overconfident predictions, potentially resulting in noise overfitting.


\begin{table}[t]
\centering
\resizebox{0.42\textwidth}{!}{
\begin{tabular}{l|c|c}
\hline 
\textbf{Avg. Error (\%)} &   \rotatebox{0}{\textbf{CIFAR100-C}} & \rotatebox{0}{\textbf{Imagenet-C}} \\\hline

\textbf{CoTTA} \cite{wang2022continual}  &32.5 & 66.8 \\ \hline
\textbf{DSS (\textit{w/} fixed $\pi$)}         &32.4&67.8  \\
\textbf{DSS (\textit{w/} fixed $\pi$\&PL)}     &32.1& 77.4 \\
\textbf{DSS (\textit{w/} fixed $\pi$\&PL\&NL)} &31.9&76.8  \\ \hline



\textbf{DSS (\textit{w/} DT)}         &31.8&66.0 \\
\textbf{DSS (\textit{w/} DT\&PL)}     &31.0&64.8 \\
\textbf{DSS (\textit{w/} DT\&PL\&NL)} &\textbf{30.9}&\textbf{64.6} \\\hline

\end{tabular}}
\caption{Adaptive threshold versus Fixed threshold. DT, PL, and NL are dynamic threshold, positive and negative learning.} \label{tab:Fixed Threshold}
\end{table}

\begin{table}[t]
\centering
\resizebox{0.38\textwidth}{!}{
\begin{tabular}{l|c|c}
\hline 
\textbf{Avg. Error (\%)} &   \rotatebox{0}{\textbf{CIFAR100-C}} & \rotatebox{0}{\textbf{Imagenet-C}} \\\hline
\textbf{CoTTA} \cite{wang2022continual}  &32.5 & 66.8 \\ \hline
\textbf{DSS ($Tp = 0.2$)}    &32.1 &   74.8   \\
\textbf{DSS ($Tp = 0.4$)}    &31.7 &   66.6   \\
\textbf{DSS ($Tp = 0.6$)}    &\textbf{30.9} & \textbf{64.6}  \\
\textbf{DSS ($Tp = 0.8$)}  &31.7 & 64.7  \\
\textbf{DSS ($Tp = 1$)} &32.0 & 64.9 \\\hline
\end{tabular}}
\caption{Temperature scaling with $Tp$ ranging from $0.2$ to $1$.} \label{tab:temperture}
\vspace{-5mm}
\end{table}

\section{Conclusion} 
In this paper, we introduce a novel method, termed Dynamic Sample Selection (DSS), based on joint positive and negative learning with a dynamic threshold for the CTDA task. Traditional methods for CTDA struggle with error accumulation since the adaptation method relies on suspected noisy pseudo-labels as a part of the adaptation process. To address this, in this paper, we consistently monitor the prediction confidence in an online manner and select low- and high-quality samples for different training strategies. To this end, we use positive learning for high-quality samples and negative learning for both low- and high-quality samples. Moreover, we have shown that our proposed method can also be applied to 3D point cloud data as well as 2D images, which showcases its versatility and potential for wide applicability. 

{\small
\bibliographystyle{ieee_fullname}
\bibliography{egbib}
}

\subfile{supp.tex}

\end{document}

%% file: supp.tex
\setcounter{section}{0}



\clearpage
\begin{center}
\textbf{Supplementary Material: \\ Continual Test-time Domain Adaptation via Dynamic Sample Selection}
\end{center}

\vspace{-8pt}
\section{Experiments}
\subsection{Imagenet-R experiment}
\noindent\textbf{ImageNet-R}~\cite{hendrycks2021many} encompasses a diverse array of shifts of ImageNet classes. These shifts include cartoons, deviant art, graffiti, embroidery, graphics, origami, paintings, patterns, plastic objects, plush objects, sculptures, sketches, tattoos, toys, and video games. The dataset comprises 200 classes and a total of 30,000 images. Here, we show the CTDA performance result of our DSS method and other baseline approaches, and experiments are conducted using the standard ResNet-50 model, pretrained on ImageNet through cross-entropy loss. In general, all baseline methods show a certain performance improvement compared to direct testing using the source model. The performance of Tent, Conjugate PL, and CoTTA methods showcases a degree of similarity, while the BN method slightly lags behind. Notably, our proposed DSS method achieves the lowest error rate of $56\%$.

\begin{table}[hb]
\centering
\resizebox{0.2\textwidth}{!}{
\begin{tabular}{l|c}
\hline 
\textbf{Method} &  \textbf{Error}\\\hline
\textbf{Source} &63.8\\
\textbf{TENT-cont} \cite{wang2021tent}    &57.3   \\
\textbf{BN Adapt \cite{li2016revisiting}} &60.3   \\
\textbf{Conjugate PL\cite{goyal2022test}} &57.3   \\
\textbf{CoTTA} \cite{wang2022continual}   &57.4   \\ \hline
\textbf{DSS (Ours)}                       &\textbf{56.0} \\ \hline 
\end{tabular}}
\vspace{-6pt}
\caption{Classification error rate (\%) on ImageNet-R~\cite{hendrycks2021many}. The best numbers are in bold.}\label{tab:ImageNet-R}
\vspace{-8pt}
\end{table}

\input{table_sup}

\vspace{-8pt}
\subsection{Modelnet40-C experiment}
\noindent\textbf{ModelNet40-C}~\cite{sun2022benchmarking} is a benchmark for assessing the robustness of our proposed method on 3D point cloud data. In this setting, 15 different forms of corruption are introduced to the original test dataset of ModelNet40~\cite{wu20153d}. For 3D experiments, random rotation and translation are used in the augmentation module to generate augmentation-weighted pseudo-labels. As shown in Table~\ref{tab:modelnet40-c}, all methods reduce error by a certain amount and DSS has the lowest error rate in average.




%% file: table_sup.tex
\begin{table*}[b!]
\centering
\resizebox{1.0\textwidth}{!}{
\begin{tabular}{l|c|c|c|c|c|c|c|c|c|c|c|c|c|c|c|c}
\hline 
\textbf{Method} & \rotatebox{0}{\textbf{uniform}}& \rotatebox{0}{\textbf{gaussian}} &\rotatebox{0}{\textbf{background}} & \rotatebox{0}{\textbf{impulse}} &\rotatebox{0}{\textbf{upsampling}} & \rotatebox{0}{\textbf{rbf}} &\rotatebox{0}{\textbf{rbf-inv}} &\rotatebox{0}{\textbf{den-dec}} &\rotatebox{0}{\textbf{dens-inc}} & \rotatebox{0}{\textbf{shear}} & \rotatebox{0}{\textbf{rot}} & \rotatebox{0}{\textbf{cut}} &\rotatebox{0}{\textbf{distort}}&\rotatebox{0}{\textbf{oclsion}} & \rotatebox{0}{\textbf{lidar}} & \textbf{Mean}\\\hline

\textbf{Source}    &14.7&18.8 &95.3&33.3&15.0&29.5&27.6&\textbf{12.9}&\textbf{10.5}&42.7&72.8&\textbf{14.9}&34.8&56.3&59.0&35.9\\
\textbf{TENT-cont} \cite{wang2021tent} &15.3&\textbf{15.6}&92.1&26.6&17.5&\textbf{26.5}&\textbf{25.1}&16.0&13.0&37.7&\textbf{58.7}&17.1&32.6&\textbf{54.1}&56.9&33.7 \\
\textbf{CoTTA} \cite{wang2022continual}&14.3&17.4&90.9&25.5&14.4&27.1&26.1&13.4&12.2&38.4&63.7&15.2&32.5&56.1&\textbf{56.6}&33.6 \\ \hline
\textbf{DSS} &\textbf{14.2}&17.7&\textbf{89.5}&\textbf{25.0}&\textbf{13.9}&26.7&25.4&13.7&12.5&\textbf{37.4}&63.6&15.4&\textbf{32.3}&54.7&58.1&\textbf{33.2}
\\ \hline 
\end{tabular}}
\vspace{-6pt}
\caption{Classification error rate (\%) on ModelNet40-C. PointNet \cite{qi2017pointnet} is adopted as the backbone. The best numbers are in bold.} 
\label{tab:modelnet40-c}
\end{table*}

%% file: PaperForReview.bbl
\begin{thebibliography}{10}\itemsep=-1pt

\bibitem{bengio2009curriculum}
Yoshua Bengio, J{\'e}r{\^o}me Louradour, Ronan Collobert, and Jason Weston.
\newblock Curriculum learning.
\newblock In {\em Proceedings of the 26th annual international conference on
  machine learning}, pages 41--48, 2009.

\bibitem{berthelot2019mixmatch}
David Berthelot, Nicholas Carlini, Ian Goodfellow, Nicolas Papernot, Avital
  Oliver, and Colin~A Raffel.
\newblock Mixmatch: A holistic approach to semi-supervised learning.
\newblock {\em Advances in neural information processing systems}, 32, 2019.

\bibitem{chen2022contrastive}
Dian Chen, Dequan Wang, Trevor Darrell, and Sayna Ebrahimi.
\newblock Contrastive test-time adaptation.
\newblock In {\em Proceedings of the IEEE/CVF Conference on Computer Vision and
  Pattern Recognition}, 2022.

\bibitem{croce2020robustbench}
Francesco Croce, Maksym Andriushchenko, Vikash Sehwag, Edoardo Debenedetti,
  Nicolas Flammarion, Mung Chiang, Prateek Mittal, and Matthias Hein.
\newblock Robustbench: a standardized adversarial robustness benchmark.
\newblock {\em arXiv preprint arXiv:2010.09670}, 2020.

\bibitem{deng2009imagenet}
Jia Deng, Wei Dong, Richard Socher, Li-Jia Li, Kai Li, and Li Fei-Fei.
\newblock Imagenet: A large-scale hierarchical image database.
\newblock In {\em 2009 IEEE conference on computer vision and pattern
  recognition}, pages 248--255. Ieee, 2009.

\bibitem{dobler2022robust}
Mario D{\"o}bler, Robert~A Marsden, and Bin Yang.
\newblock Robust mean teacher for continual and gradual test-time adaptation.
\newblock {\em arXiv preprint arXiv:2211.13081}, 2022.

\bibitem{gan2022decorate}
Yulu Gan, Xianzheng Ma, Yihang Lou, Yan Bai, Renrui Zhang, Nian Shi, and Lin
  Luo.
\newblock Decorate the newcomers: Visual domain prompt for continual test time
  adaptation.
\newblock {\em arXiv preprint arXiv:2212.04145}, 2022.

\bibitem{gandelsman2022test}
Yossi Gandelsman, Yu Sun, Xinlei Chen, and Alexei~A Efros.
\newblock Test-time training with masked autoencoders.
\newblock {\em arXiv preprint arXiv:2209.07522}, 2022.

\bibitem{goyal2022test}
Sachin Goyal, Mingjie Sun, Aditi Raghunathan, and J~Zico Kolter.
\newblock Test time adaptation via conjugate pseudo-labels.
\newblock {\em Advances in Neural Information Processing Systems},
  35:6204--6218, 2022.

\bibitem{hendrycks2021many}
Dan Hendrycks, Steven Basart, Norman Mu, Saurav Kadavath, Frank Wang, Evan
  Dorundo, Rahul Desai, Tyler Zhu, Samyak Parajuli, Mike Guo, et~al.
\newblock The many faces of robustness: A critical analysis of
  out-of-distribution generalization.
\newblock In {\em Proceedings of the IEEE/CVF International Conference on
  Computer Vision}, pages 8340--8349, 2021.

\bibitem{hendrycks2018benchmarking}
Dan Hendrycks and Thomas Dietterich.
\newblock Benchmarking neural network robustness to common corruptions and
  perturbations.
\newblock In {\em International Conference on Learning Representations}, 2019.

\bibitem{hendrycks2019benchmarking}
Dan Hendrycks and Thomas Dietterich.
\newblock Benchmarking neural network robustness to common corruptions and
  perturbations.
\newblock {\em arXiv preprint arXiv:1903.12261}, 2019.

\bibitem{hong2023pointcam}
Jie Hong, Shi Qiu, Weihao Li, Saeed Anwar, Mehrtash Harandi, Nick Barnes, and
  Lars Petersson.
\newblock Pointcam: Cut-and-mix for open-set point cloud learning.
\newblock {\em arXiv preprint arXiv:2212.02011}, 2023.

\bibitem{kim2019nlnl}
Youngdong Kim, Junho Yim, Juseung Yun, and Junmo Kim.
\newblock Nlnl: Negative learning for noisy labels.
\newblock In {\em Proceedings of the IEEE International Conference on Computer
  Vision}, pages 101--110, 2019.

\bibitem{kim2021joint}
Y. Kim, J. Yun, H. Shon, and J. Kim.
\newblock Joint negative and positive learning for noisy labels.
\newblock In {\em 2021 IEEE/CVF Conference on Computer Vision and Pattern
  Recognition (CVPR)}, pages 9437--9446, Los Alamitos, CA, USA, jun 2021. IEEE
  Computer Society.

\bibitem{krizhevsky2009learning}
Alex Krizhevsky, Geoffrey Hinton, et~al.
\newblock Learning multiple layers of features from tiny images.
\newblock 2009.

\bibitem{li2016revisiting}
Yanghao Li, Naiyan Wang, Jianping Shi, Jiaying Liu, and Xiaodi Hou.
\newblock Revisiting batch normalization for practical domain adaptation.
\newblock {\em arXiv preprint arXiv:1603.04779}, 2016.

\bibitem{liang2020we}
Jian Liang, Dapeng Hu, and Jiashi Feng.
\newblock Do we really need to access the source data? source hypothesis
  transfer for unsupervised domain adaptation.
\newblock In {\em International Conference on Machine Learning}, pages
  6028--6039, 2020.

\bibitem{niu2022efficient}
Shuaicheng Niu, Jiaxiang Wu, Yifan Zhang, Yaofo Chen, Shijian Zheng, Peilin
  Zhao, and Mingkui Tan.
\newblock Efficient test-time model adaptation without forgetting.
\newblock {\em arXiv preprint arXiv:2204.02610}, 2022.

\bibitem{qi2017pointnet}
Charles~R Qi, Hao Su, Kaichun Mo, and Leonidas~J Guibas.
\newblock Pointnet: Deep learning on point sets for 3d classification and
  segmentation.
\newblock In {\em Proceedings of the IEEE conference on computer vision and
  pattern recognition}, pages 652--660, 2017.

\bibitem{rizve2021defense}
Mamshad~Nayeem Rizve, Kevin Duarte, Yogesh~S Rawat, and Mubarak Shah.
\newblock In defense of pseudo-labeling: An uncertainty-aware pseudo-label
  selection framework for semi-supervised learning.
\newblock {\em arXiv preprint arXiv:2101.06329}, 2021.

\bibitem{sohn2020fixmatch}
Kihyuk Sohn, David Berthelot, Nicholas Carlini, Zizhao Zhang, Han Zhang,
  Colin~A Raffel, Ekin~Dogus Cubuk, Alexey Kurakin, and Chun-Liang Li.
\newblock Fixmatch: Simplifying semi-supervised learning with consistency and
  confidence.
\newblock {\em Advances in neural information processing systems}, 33:596--608,
  2020.

\bibitem{sun2022benchmarking}
Jiachen Sun, Qingzhao Zhang, Bhavya Kailkhura, Zhiding Yu, Chaowei Xiao, and
  Z~Morley Mao.
\newblock Benchmarking robustness of 3d point cloud recognition against common
  corruptions.
\newblock {\em arXiv preprint arXiv:2201.12296}, 2022.

\bibitem{sun2020test}
Yu Sun, Xiaolong Wang, Zhuang Liu, John Miller, Alexei Efros, and Moritz Hardt.
\newblock Test-time training with self-supervision for generalization under
  distribution shifts.
\newblock In {\em International conference on machine learning}, pages
  9229--9248, 2020.

\bibitem{NIPS2017_68053af2}
Antti Tarvainen and Harri Valpola.
\newblock Mean teachers are better role models: Weight-averaged consistency
  targets improve semi-supervised deep learning results.
\newblock {\em Advances in neural information processing systems}, 30, 2017.

\bibitem{uy-scanobjectnn-iccv19}
Mikaela~Angelina Uy, Quang-Hieu Pham, Binh-Son Hua, Duc~Thanh Nguyen, and
  Sai-Kit Yeung.
\newblock Revisiting point cloud classification: A new benchmark dataset and
  classification model on real-world data.
\newblock In {\em International Conference on Computer Vision (ICCV)}, 2019.

\bibitem{wang2021tent}
Dequan Wang, Evan Shelhamer, Shaoteng Liu, Bruno Olshausen, and Trevor Darrell.
\newblock Tent: Fully test-time adaptation by entropy minimization.
\newblock In {\em International Conference on Learning Representations}, 2021.

\bibitem{wang2022continual}
Qin Wang, Olga Fink, Luc Van~Gool, and Dengxin Dai.
\newblock Continual test-time domain adaptation.
\newblock In {\em Proceedings of the IEEE/CVF Conference on Computer Vision and
  Pattern Recognition}, pages 7201--7211, 2022.

\bibitem{wang2022freematch}
Yidong Wang, Hao Chen, Qiang Heng, Wenxin Hou, Marios Savvides, Takahiro
  Shinozaki, Bhiksha Raj, Zhen Wu, and Jindong Wang.
\newblock Freematch: Self-adaptive thresholding for semi-supervised learning.
\newblock {\em arXiv preprint arXiv:2205.07246}, 2022.

\bibitem{wang2019dynamic}
Yue Wang, Yongbin Sun, Ziwei Liu, Sanjay~E Sarma, Michael~M Bronstein, and
  Justin~M Solomon.
\newblock Dynamic graph cnn for learning on point clouds.
\newblock {\em Acm Transactions On Graphics (tog)}, 38(5):1--12, 2019.

\bibitem{wu20153d}
Zhirong Wu, Shuran Song, Aditya Khosla, Fisher Yu, Linguang Zhang, Xiaoou Tang,
  and Jianxiong Xiao.
\newblock 3d shapenets: A deep representation for volumetric shapes.
\newblock In {\em Proceedings of the IEEE conference on computer vision and
  pattern recognition}, pages 1912--1920, 2015.

\bibitem{yin2019fourier}
Dong Yin, Raphael Gontijo~Lopes, Jon Shlens, Ekin~Dogus Cubuk, and Justin
  Gilmer.
\newblock A fourier perspective on model robustness in computer vision.
\newblock {\em Advances in Neural Information Processing Systems}, 32, 2019.

\bibitem{BMVC2016_87}
Sergey Zagoruyko and Nikos Komodakis.
\newblock Wide residual networks.
\newblock In Edwin R.~Hancock Richard C.~Wilson and William A.~P. Smith,
  editors, {\em Proceedings of the British Machine Vision Conference (BMVC)},
  pages 87.1--87.12. BMVA Press, September 2016.

\bibitem{zhang2021flexmatch}
Bowen Zhang, Yidong Wang, Wenxin Hou, Hao Wu, Jindong Wang, Manabu Okumura, and
  Takahiro Shinozaki.
\newblock Flexmatch: Boosting semi-supervised learning with curriculum pseudo
  labeling.
\newblock {\em Advances in Neural Information Processing Systems},
  34:18408--18419, 2021.

\end{thebibliography}
